\documentclass[]{unifiedreward}
\usepackage{fix-cm}
\usepackage{helvet}           

\usepackage{nicefrac}         
\usepackage{siunitx}         

\usepackage{longtable}
\usepackage{booktabs}

\usepackage{tabularx}         
\usepackage{array}           
\usepackage{makecell}         
\usepackage{threeparttable}  
\usepackage{xspace}

\usepackage{wrapfig}          
\usepackage{float}            
\usepackage[export]{adjustbox}

\usepackage{tikz}             
\usepackage{pgfplots}         
\usepackage{pgf-pie}

\usepackage{listings}        
\usepackage{ragged2e}        
\usepackage{comment}

\usepackage{pifont}           
\usepackage{fontawesome}

\usepackage{enumitem}        
\usepackage{titletoc}         
\usepackage{minitoc}          
\usepackage[toc,page,header]{appendix}

\usepackage{url}              
\usepackage[hang,flushmargin]{footmisc}  
\pgfplotsset{compat=1.18}

\usepackage{amsmath} 
\newcommand{\eg}{\textit{e.g.}\xspace} 
\newcommand{\ie}{\textit{i.e.}\xspace}

\definecolor{lightblue}{RGB}{200, 230, 255}  
\definecolor{headerblue}{RGB}{150, 200, 255}

\newcounter{examplebox}

\makeatletter
\newcommand\blfootnote[1]{%
  \begingroup
  \renewcommand\thefootnote{}\footnote{#1}%
  \addtocounter{footnote}{-1}%
  \endgroup
}
\makeatother

\title{
   \textit{DomainShuttle}: Freeform Open Domain Subject-driven Text-to-video Generation
}

\author{
    Nan Chen\textsuperscript{*}, 
    Yiyang Cai\textsuperscript{*},   
    Weinan Jia, \\
    Zhuowei Chen ,
    Wen Zhou,
     Wenhan Luo\textsuperscript{$\dagger$}
}

\affiliation[1]{\mbox{Hong Kong University of Science and Technology}}

\abstract{  Open domain subject-driven text-to-video (S2V) generation has drawn significant interest in academia and industry. Open domain S2V mainly involves two scenarios: in-domain, which requires retaining the reference subject features as much as possible, and cross-domain, which preserves the intrinsic features of the subject while allowing subject-irrelevant properties to vary flexibly according to the text prompt. Existing methods primarily focus on maximizing subject fidelity in in-domain scenarios, which limits their editability and adaptability in cross-domain scenarios, such as novel styles, semantic combinations, or domain attributes. In this study, we propose that an ideal S2V method should flexibly shuttle between different domains, achieving strong performance in both in-domain and cross-domain scenarios. To this end, we propose \textit{DomainShuttle}, which could achieve high fidelity and generative flexibility for open domain video personalization. Specifically, we introduce Domain-MoT, which decouples videos and reference features and introduces the domain-aware AdaLN for domain-specific modeling of reference images. We then introduce the Video-Reference DualRoPE scheme, which places reference image tokens and video tokens in separate RoPE spaces to enable precise subject-level spatial modeling, and Cross-Pair Consistent Loss, which aims to extract intrinsic subject features unaffected by irrelevant features. Extensive experiments demonstrate that \textit{DomainShuttle} achieves significant performance improvements over existing methods, exhibiting high subject fidelity and generative flexibility across diverse open domain application scenarios.
}

\checkdata[Website]{\url{https://cn-makers.github.io/DomainShuttle/}}
\checkdata[Code]{\url{https://github.com/HKUST-C4G/DomainShuttle}}

\begin{document}
\maketitle

\blfootnote{$^\dagger$ Corresponding Authors.}
\blfootnote{\textsuperscript{*} Equal Contribution.}

\section{Introduction}

\begin{figure*}[t]
    \includegraphics[width=0.98\linewidth]{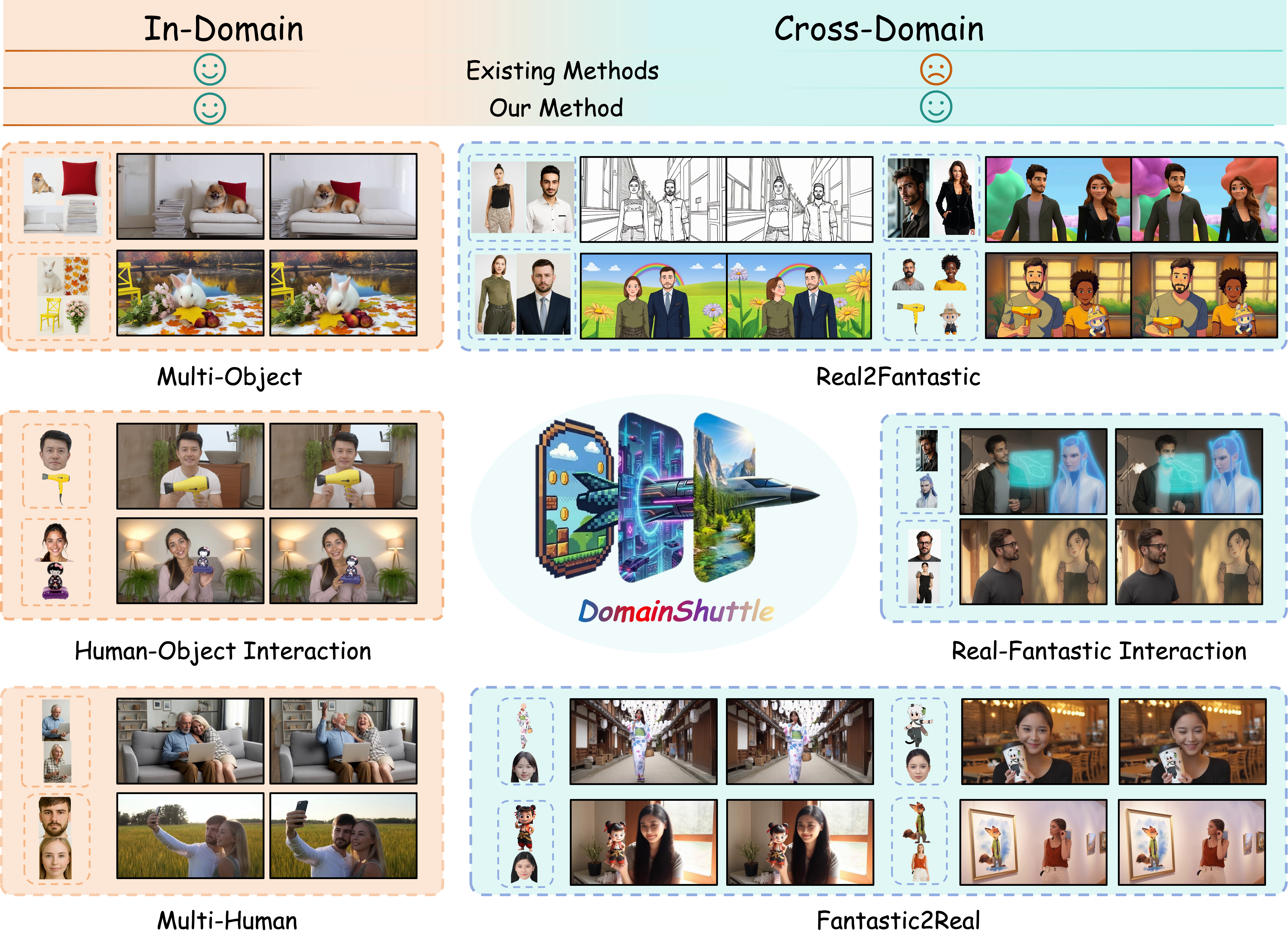}
    \caption{\textit{DomainShuttle} demonstrates strong performance in both in-domain and cross-domain scenarios. Existing S2V methods typically focus on in-domain scenarios, with limited effectiveness in cross-domain scenarios. In contrast, \textit{DomainShuttle} can achieve high subject consistency and flexible generation in both scenarios.}
    \label{fig:teaser}
\end{figure*}

Subject-driven video generation (S2V) is an important task in video generation, with broad applications in advertising, creative design, and AI filmmaking. These diverse applications require S2V to acquire ``open-domain'' capabilities, which can precisely extract subject features from reference images and flexibly generate videos regardless of domain types. Specifically, given reference images of subjects  (\eg, humans, objects, fantasy IPs, and backgrounds), the open-domain S2V model can retain the original appearance of these subjects in in-domain videos, as shown on the left of Fig. \ref{fig:teaser}, or flexibly transform them into cross-domain videos while preserving their intrinsic subject features, as shown on the right of Fig. \ref{fig:teaser}. The key challenge of open-domain S2V lies in how to achieve flexible and consistent subject generation across both in-domain and cross-domain scenarios while precisely preserving their distinctive subject features, rather than just mimicking reference images.

Existing S2V studies primarily aim to improve subject fidelity in in-domain scenarios, \ie, preserving as many subject features as possible without altering their attributes or style. In-domain examples are shown on the left of Fig. \ref{fig:teaser}. Early studies \cite{ID-Animator, ConsisID, PersonalVideo, MagicID} investigate identity preservation for single-subject S2V. With the emergence of numerous personalized datasets, recent studies have shifted towards multi-subject S2V. Phantom \cite{phantom} and VACE \cite{vace} design novel reference feature injection schemes to improve multi-subject fidelity. Some methods \cite{chen2025humo,deng2026magref,chen2025first} leverage image-to-video (I2V) models for strong subject fidelity priors. Additionally, some studies \cite{li2026bindweave,hunyuancustom,deng2025cinema} utilize multimodal large language models to enhance subject fidelity in in-domain scenarios. In summary, existing methods mainly focus on improving subject fidelity in in-domain scenarios.

However, existing methods usually neglect more diverse and creative cross-domain scenarios. Such cross-domain scenarios involve generating real-world subjects in different fantasy domains, mapping fantasy domain subjects to various real-world objects, and constructing complex interactions between real-world subjects and fantastic subjects, as shown on the right of Fig. \ref{fig:teaser}. Restricting modeling to in-domain scenarios sacrifices a certain degree of editability and flexibility in complex scenarios, making it difficult for models to simultaneously preserve subject consistency and flexible adaptability to new styles, semantic combinations, and domain attributes in cross-domain scenarios. As a result, the generalization and creative potential of existing methods in open domain scenarios are limited to some extent.

We propose that an ideal S2V method could freely \textbf{shuttle} between different domains, which should achieve a dual optimization of high fidelity of the reference subject and generation flexibility in open-domain scenarios (including in-domain and cross-domain). Specifically, subject features of the reference images should only affect intrinsic subject attributes in the generated video (\eg, hairstyle, skin color, and clothing), while the subject-irrelevant features (\eg, lighting, style, and domain attributes) should flexibly adapt according to the text instructions.

Based on the above analysis, we propose a novel S2V framework, \textit{\textbf{DomainShuttle}}, which achieves joint optimization of subject consistency and generation flexibility by designing independent information processing paths for video and reference branches, along with additional domain-aware modeling in the reference image branch. \textit{DomainShuttle} includes three key components: (1) \textbf{DomainMoT} (Mixture-of-Transformers), which aims to decouple videos and reference features to facilitate domain-aware reference modeling; (2) \textbf{Video-Reference DualRoPE}, which assigns reference image tokens to a separate RoPE space from video tokens, enabling precise subject-level spatial distance relationships; and (3) \textbf{Cross-Pair Consistent Loss}, which aims to extract intrinsic subject features unaffected by irrelevant features. Specifically, DomainMoT introduces an independent attention mapping pathway for reference image features within the in-context self-attention to effectively disentangle them from video features, and further introduces a domain-aware AdaLN to distinguish different domains in the reference feature space. Furthermore, Video-Reference DualRoPE assigns reference image tokens to a separate RoPE space independent of the video, enabling precise subject-level control by explicitly separating different subjects in the latent space while pulling representations of the same subject closer together. Cross-Pair Consistent Loss aligns two sets of reference images corresponding to the same video to better capture the intrinsic subject features, which are unaffected by irrelevant features such as lighting, composition, or style. We conduct comprehensive evaluations in both in-domain and cross-domain scenarios using video quality, text alignment, and subject consistency metrics. The experimental results show that \textit{DomainShuttle} achieves competitive performance, with a significant 18.7\% improvement in Cross-Domain Score over the SOTA methods.

Our main contributions are summarized as follows:
\begin{itemize}[label=\textbullet]
    \item  We propose a novel S2V framework termed \textit{DomainShuttle}, which decouples video and reference images into independent branches via the DomainMoT module. Within DomainMoT, the domain-aware AdaLN is introduced to facilitate domain-specific modeling of reference images, thereby enabling high subject fidelity and flexible controllability in open-domain scenarios.
    \item   Building on DomainMoT, we further design the Video-Reference DualRoPE mechanism to accurately distinguish subject-level spatial distance, and Cross-Pair Consistent Loss to precisely extract the intrinsic subject features.
    \item   Extensive experiments show that \textit{DomainShuttle} comprehensively outperforms existing methods in terms of subject consistency and text controllability across various complex scenarios, especially achieving an 18.7\% improvement in Cross-Domain Score over the SOTA methods. 

\end{itemize}
\section{Related Work}

\subsection{Video Foundation Model}

Video diffusion models have attracted substantial attention in academia and industry in recent years, profoundly advancing video creation. Early approaches \cite{ho2022video, svd,guo2024animatediff} are mainly based on UNet, augmenting image diffusion models with temporal modules to synthesize videos. With the advent of the DiT architecture \cite{DiT}, visual generative models have scaled in both capacity and capability. Currently, an increasing number of video diffusion models adopt DiT to further improve generation quality and controllability, such as CogVideoX \cite{yang2024cogvideox}, HuanyuanVideo \cite{kong2024hunyuanvideo}, Seedance \cite{chen2025seedance}, and the Wan series \cite{wan2025wan}. As the base models of video diffusion improve, application scenarios\cite{phantom,hunyuancustom,chen2025humo,ye2026unified,ye2026visual,cai2026foundation,ye2025stylemaster,yang2025infinitetalk,ditto1m,insvie,reco} for video generation are becoming increasingly diverse, such as subject-driven video generation \cite{phantom,hunyuancustom,chen2025humo}, audio-visual generation \cite{chen2025humo,yang2025infinitetalk}, and reference video-based video editing \cite{ditto1m,insvie,reco}.

\subsection{Subject-driven Video Generation}

Subject-driven text-to-video generation aims to synthesize videos based on user-provided reference images by preserving specified subject features (\eg, identity, domain semantics, style, and attribute features) under textual guidance. Early studies \cite{ID-Animator, ConsisID, PersonalVideo, MagicID} primarily focus on human-centered single-identity generation. Recent research \cite{phantom,chen2025humo,deng2026magref,chen2025first,li2026bindweave,hunyuancustom,fei2025skyreels,deng2025cinema, ConceptMaster, VideoAlchemist} has shifted towards more general scenarios, focusing on multi-subject video personalization (\eg, human, object, and background). Phantom \cite{phantom} proposes a dynamic injection scheme to support the generation of multiple subjects. Some studies \cite{chen2025humo,deng2026magref,chen2025first} leverage the inherent subject-preserving capabilities of I2V models to achieve multi-subject preservation more quickly, but this approach often suffers from copy-paste issues. Several studies \cite{li2026bindweave,hunyuancustom,deng2025cinema} incorporate features from multimodal large language models (MLLM) to enhance the understanding of spatiotemporal instructions. However, existing methods only consider maintaining identity fidelity while neglecting the preservation of the inherent flexible cross-domain capabilities of base models, which weakens the flexibility and creative potential in complex open-domain scenarios. Based on this, we seek to effectively preserve open-domain generation capabilities while ensuring high identity fidelity in this work.

\begin{figure*}[t]
    \includegraphics[width=0.98\linewidth]{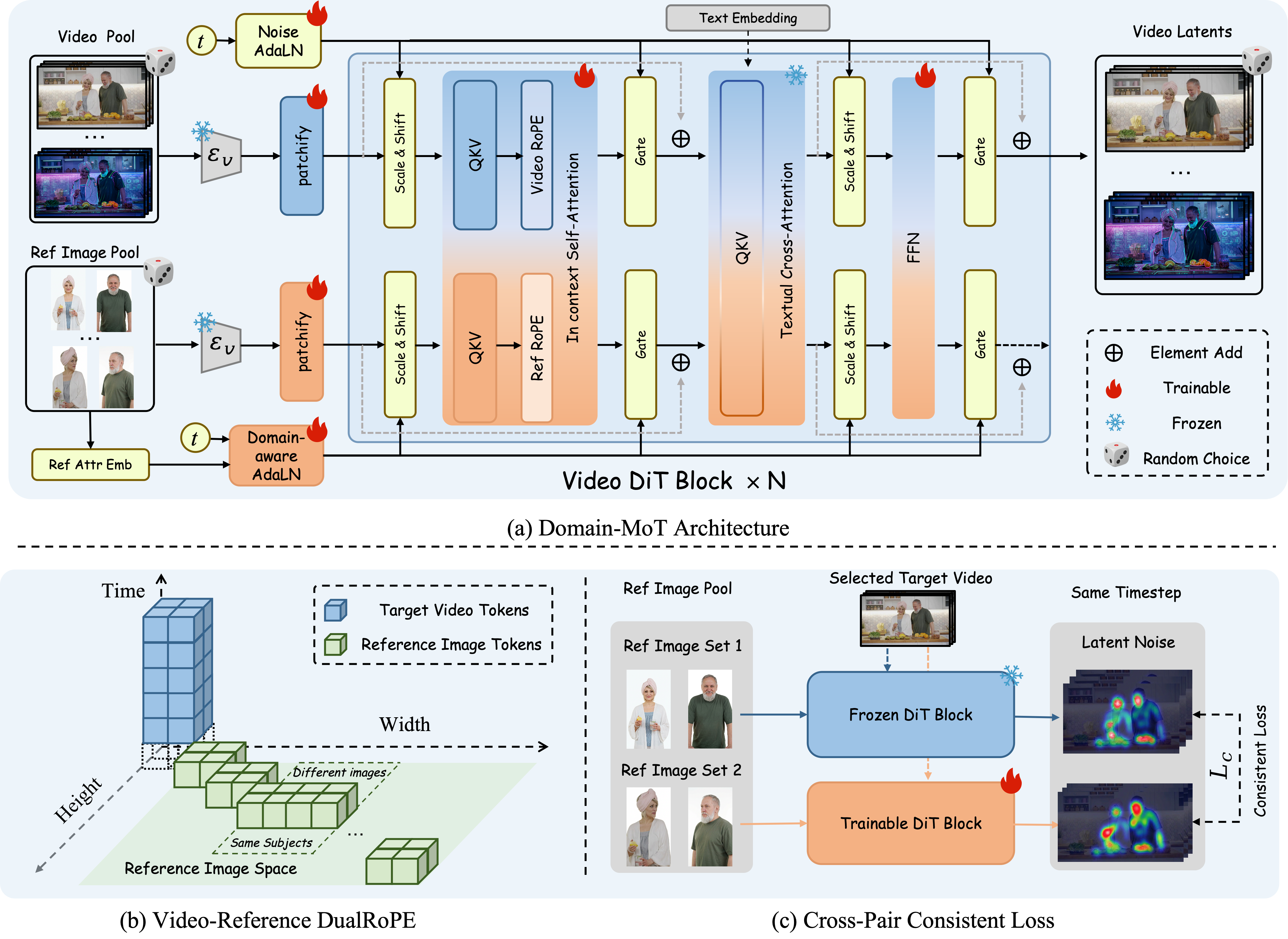}
    \caption{Overview of \textit{DomainShuttle}. (a) The reference images and videos are injected into the decoupled branches of Domain-MoT, facilitating domain-aware AdaLN guidance in the reference branch. (b)  Video-Reference DualRoPE disentangles the RoPE spaces of reference images and videos for precise subject-level spatial distance relationships. (c) Cross-Pair Consistent Loss aligns the features of two sets of reference images, enabling the model to learn intrinsic subject features rather than redundant features.}
    \label{fig:method_overview}
\end{figure*}
\section{Methodology}

\textit{DomainShuttle} is designed to advance flexible and high-fidelity cross-domain personalization, which consists of three modules: Domain-MoT, Video-Reference DualRoPE, and Cross-Pair Consistent Loss, as illustrated in Fig. \ref{fig:method_overview}. During training,  video latents and reference image features are processed through separate branches in \textbf{DomainMoT} to achieve feature disentanglement, facilitating the incorporation of domain-aware AdaLN modulation in the reference branch. We then introduce \textbf{Video-Reference DualRoPE}, which assigns reference image tokens to a RoPE space independent of video tokens, enabling precise subject-level control: different subjects are explicitly separated, while representations of the same subject are pulled closer in the latent space. Finally, the \textbf{Cross-Pair Consistent Loss} (CCL) aligns two sets of reference images corresponding to the same video to precisely capture subject-specific features. In this section, we first introduce the preliminaries of video generation models. We then present the design of DomainMoT, Video-Reference  DualRoPE, and the Cross-Pair Consistent Loss in detail. Finally, we describe the construction of our training dataset.

\subsection{Preliminaries} 

Our method is trained on the DiT-based video generation model \cite{wan2025wan}. During training, the text encoder  $\boldsymbol{\mathcal{E}}_t(\cdot)$ \cite{chung2023unimax} encodes the text into text feature $\boldsymbol{c}_t$, and the 3D VAE encoder  $\boldsymbol{\mathcal{E}}_v(\cdot)$ encodes the video and $N$ reference images $\mathcal{R} = \{ I_0, I_1, \cdots, I_{N-1} \}$ into video latents $\boldsymbol{z}_1$ and reference image features $\boldsymbol{c}_r$. The model is optimized by flow-matching loss \cite{lipman2023flow}, which is defined as follows:

\begin{equation}
\mathcal{L}_{\mathrm{FM}}=\mathbb{E}_{t,\boldsymbol{z}_0,\boldsymbol{z}_1}\|{G}_\theta(\boldsymbol{z}_t,t,\boldsymbol{c}_t,\boldsymbol{c}_r)-(\boldsymbol{z}_1-\boldsymbol{z}_0)\|_2^2,
\label{eq:flow_matching_loss}
\end{equation}
where continuous time $t\in[0,1]$. $\boldsymbol{z}_0$ denotes a sample drawn from the prior distribution, and $\boldsymbol{z}_1$ denotes video latent encoded by 3D VAE. ${G}_\theta$ represents a learnable vector field parameterized by $\theta$.

\subsection{Model Architecture} 
\subsubsection{Domain-MoT}

A key issue in cross-domain S2V is the entanglement between intrinsic subject features and domain-specific attributes, which makes it difficult to preserve subject features while enabling flexible transitions across domains. To address this issue, we propose Domain-MoT, as shown in Fig. \ref{fig:method_overview} (a). Specifically, Domain-MoT decouples video latents and reference image features into two independent processing paths, and explicitly injects domain attributes into the reference image branch via Domain-aware AdaLN to distinguish different domains in the feature space, achieving more precise cross-domain S2V.

As shown in Fig. \ref{fig:method_overview} (a), the video latents and reference image features after 3D VAE encoding are separately patchified to extract their patch embeddings $\boldsymbol{f}_v$ and $\boldsymbol{f}_r$, which facilitates the subsequent integration of the in-context self-attention and Domain-aware AdaLN, achieving more stable reference injection.

Next, Domain-MoT employs in-context self-attention with independent QKV projections and independent RoPE for video latents and reference features. This decoupled design preserves the inherent capability of the video branch as the base model, while allowing the reference image branch to focus on extracting more precise subject features for flexible and high-fidelity personalization. The in-context self-attention is shown below:

\begin{equation}
\label{eq:attention}
 \text{Softmax}\left( \frac{ [ R_{v}(  \boldsymbol{Q}_{v}); R_{r}(\boldsymbol{Q}_{r})] \cdot [ R_{v}(  \boldsymbol{K}_{v}); R_{r}(\boldsymbol{K}_{r})]}{\sqrt{d}} \right)  [\boldsymbol{V}_{v},\boldsymbol{V}_{r}], 
\end{equation}
Where $ R_{v}$ and $ R_{r}$ denote the RoPE applied to the video branch and reference image branch, respectively, which will be discussed in the next subsection. The queries $\boldsymbol{Q}_v=\boldsymbol{W}^q_v\cdot \boldsymbol{f}_v$ and $\boldsymbol{Q}_r=\boldsymbol{W}^q_r \cdot \boldsymbol{f}_r$,  keys  $\boldsymbol{K}_v=\boldsymbol{W}^k_v \cdot \boldsymbol{f}_v$ and $\boldsymbol{K}_r=\boldsymbol{W}^k_r \cdot \boldsymbol{f}_r$, values $\boldsymbol{V}_v=\boldsymbol{W}_v^v \cdot \boldsymbol{f}_v$ and $\boldsymbol{V}_r=\boldsymbol{W}^v_r \cdot f_r$. $\boldsymbol{W}^q$,  $\boldsymbol{W}^k$ and $\boldsymbol{W}^v$ are weight parameters. \( [; ] \) denotes feature concatenation. 

To preserve intrinsic text guidance capability of the base model across complex cross-domain scenarios, we freeze the textual cross-attention during training. Textual cross attention enables the interaction between textual features $\boldsymbol{f}_t$  and concatenated visual features $\boldsymbol{f}_c = [\boldsymbol{f}_v; \boldsymbol{f}_r]$, as follows:
\begin{equation}
\label{eq:attention2}
 \text{Softmax}\left( \frac{   \boldsymbol{Q}_{c} \cdot \boldsymbol{K}_{t}}{\sqrt{d}} \right)  \boldsymbol{V}_{t}, 
\end{equation}
where the query $\boldsymbol{Q}_c=\boldsymbol{W}^q\cdot \boldsymbol{f}_c$,  key  $\boldsymbol{K}_t=\boldsymbol{W}^k \cdot \boldsymbol{f}_t$ and value $\boldsymbol{V}_t=\boldsymbol{W}^v \cdot \boldsymbol{f}_t$.

\paragraph{\textbf{Domain-aware AdaLN.}} We then introduce Domain-aware AdaLN, a novel mechanism that aims to achieve flexible cross-domain generation by decoupling domain attributes. Existing S2V methods typically modulate reference tokens and video latents indiscriminately, causing domain attributes from the reference image to be entangled with video latents. Domain-aware AdaLN addresses this by structurally decoupling the noise AdaLN and reference AdaLN. Notably, the reference AdaLN is modulated by both the reference domain attributes and time features, while the noise AdaLN is modulated only by time features.

The explicit injection mechanism of reference domain attributes decouples features of content and domain, enabling cross-domain generation without disturbing the content and temporal structure by simply exchanging domain attributes. On the one hand, in in-domain generation scenarios, this mechanism can utilize in-domain prior knowledge to improve generation quality; on the other hand, in cross-domain scenarios, replacing the injected domain attributes achieves better cross-domain generation results. Specifically, the AdaLN mechanisms for video latent and reference image features are as follows:

\begin{equation}
\left\{
\begin{aligned}
& \hat{\boldsymbol{f}}_v \, =  \, g_{v}(t) \,\odot\, \big[ \mathcal{F}\big( \mathrm{LN}( \boldsymbol{f}_v) \,\odot\, (1+\gamma_{v}(t))  \;+\; \beta_{v}(t) \big)\big] + \boldsymbol{f}_v , \\
& \hat{\boldsymbol{f}}_r \, = \, g_{r}(t,a) \,\odot\, \big[ \mathcal{F} \big( \mathrm{LN}( \boldsymbol{f}_r) \,\odot\,  (1+\gamma_{r}(t,a))  \;+\; \beta_{r}(t,a)\big) \big] +\boldsymbol{f}_r  ,
\end{aligned}
\right.
\end{equation}
where $t$ denotes time features and $a \in  \{A_1, A_2, A_3, \ldots, A_K\}$ denotes one of the $K$ domain attributes. $\mathrm{LN}$ denotes layer normalization and  $\odot$ denotes Hadamard product.  $\hat{\boldsymbol{f}}_v$ and $\hat{\boldsymbol{f}}_r$ denote the modulated video features and reference features, respectively. The modulation coefficients are given by the scale $\gamma \in \mathbf{R}^{d}$, shift $\beta \in\mathbf{R}^{d}$, and residual gate $g \in \mathbf{R}^{d}$ conditioned on $t$ and $a$. $\mathcal{F}(\cdot)$ denotes general residual functions (\eg, attention and FFN).

\subsubsection{ Video-Reference DualRoPE}
\label{3.2.2}

Currently, DiT-based video generation models \cite{wan2025wan,yang2024cogvideox} commonly adopt Rotary Positional Encoding (RoPE) to distinguish positional information among video tokens. RoPE assigns a positional index to each token, modulating the interaction strength between different tokens, which means that tokens with closer indices correspond to shorter latent distances. Specifically, each video token is assigned a positional index $(i,j,k)$, where $i\in[0,f-1]$, $j\in[0,h-1]$, and $k\in[0,w-1]$. Here, $f$, $h$, and $w$ denote the number of frames, height, and width in the video latent space, respectively.

Existing methods mainly inherit the RoPE mechanism of the base model, applying the reference image RoPE by treating each reference image as an additional video frame. This mechanism distinguishes different reference images solely through temporal indices, ignoring that multiple reference subjects lack temporal continuity and that multiple reference images may jointly describe a single subject. Notably, a reference subject is not equivalent to a single reference image, as one subject can correspond to multiple reference images. To explicitly disentangle video tokens from different reference subjects, we propose \textbf{Video-Reference (VR) DualRoPE}, which allocates reference image tokens into a RoPE space fully decoupled from the video token space, enabling precise RoPE spatial distance relationships among different reference subjects, as shown in Fig. \ref{fig:method_overview} (b). Through this design, the model achieves joint optimization of subject consistency and text controllability in open-domain scenarios. The video noise RoPE $R_v(i,j,k)$ and the reference image RoPE $R_r(i,j,k)$ are defined as follows:
\begin{equation}
\left\{
\begin{aligned}
& R_v(i,j,k) = \theta (i+1,j,k)  , \\
&R_r(i,j,k) = \theta(0,j+h\times(m+1),k+w\times(n+1)),
\end{aligned}
\right.
\end{equation}
where $m \in [0, M-1] $ denotes the $m$-th reference subject, $n \in [0, N-1]$ denotes the $n$-th reference image, and $\theta$ represents the rotation function. For reference images, the temporal index is set to 0 while the temporal index for video starts from 1, explicitly separating the reference RoPE space from the video RoPE space. For two adjacent reference images representing different subjects, the RoPE offset is $\Delta=(0,h,w)$. When two reference images represent different parts of the same subject, they are treated as sub-images of a large reference image, whose offset is set to $\Delta=(0,0,w)$. This design distinguishes semantic differences between different reference images and keeps images of the same subject closer in the latent space, thus explicitly establishing their identity associations.

\subsubsection{ Cross-Pair Consistent Loss}

To further enhance the flexibility of our model in various complex cross-domain scenarios, we propose a \textbf{Cross-pair Consistent Loss} (CCL) to precisely extract key features from references.

We construct multiple sets of reference images for each video, forming a reference pool (see Fig. \ref{data_pipeline} for dataset details). During training, two sets of reference images are randomly sampled from the pool, and the generated video latent noises are aligned at the same timestep as shown in Fig. \ref{fig:method_overview} (c). This encourages the model to extract shared features (\eg, shape, texture style, and subject identity) from different references and suppresses overfitting to redundant features in a single frame. Different reference sets vary in viewpoint, occlusion, motion blur, and illumination. By forcing a learnable branch to match a frozen reference branch, the model learns representations that are insensitive to these perturbations, thereby improving flexibility in complex cross-domain scenarios. Compared to randomly sampling a single reference set at different timesteps during training, our strategy achieves more precise consistency alignment at the same noise level, leading to more effective extraction of precise subject features in the reference images. The Cross-Pair Consistent Loss $\mathcal{L}_{\mathrm{C}}$ is defined as:

\begin{equation}
\mathcal{L}_{\mathrm{C}}=\|{G}_\theta(\boldsymbol{z}_t,t,\boldsymbol{c}_t,\boldsymbol{c}_r)-{G}^*_\theta(\boldsymbol{z}_t,t,\boldsymbol{c}_t,\boldsymbol{c}^*_r)\|_2^2,
\label{eq:loss_2}
\end{equation}
where $\boldsymbol{c}_r$ and $\boldsymbol{c}^*_r$ represent two different sets of reference image features. The ${G}^*_\theta$ branch is frozen, while the ${G}_\theta$ branch is trainable.

\subsection{Training Data Pipeline}
\label{data_pipeline}
Our training data comprises two components: image and video personalization datasets. We build a 200K image dataset containing multiple subjects based on open-source datasets \cite{UNO,echo-4o} to endow the model with basic personalization capability. For video personalization, we use three datasets: Phantom-Data \cite{phantom-data}, OpenS2V \cite{yuan2025opens2v}, and Ditto-1M \cite{zhang2025region}. We first filter low-quality videos using aesthetics and motion metrics. Phantom-Data naturally provides numerous cross-pairs after retrieval, enabling direct application of Cross-Pair Consistent Loss on its cross-pair set. For OpenS2V and Ditto-1M, we utilize Grounding-DINO \cite{groundingdino} for object detection and SAM2 \cite{sam2} for multi-frame segmentation to build a reference image set for Cross-Pair Consistent Loss. We then use visual-semantic alignment via MLLM \cite{qwen25} to remove low-quality references (\eg, incomplete, blurred, or semantically mismatched subjects). Notably, Ditto-1M is a video editing dataset in which segmented reference images can align with multiple edited and source videos, facilitating ``single reference set → multiple videos'' pair construction, which further promotes the extraction of precise subject features. 

In total, we obtain a 750K high-quality, open-domain video personalization dataset covering various scenarios (\eg, humans, objects, fantasy subjects, and backgrounds) and supporting cross-pair configurations of both ``multiple reference set → single video'' and ``single reference set → multiple videos''.  In the training dataset, only 50K from the Ditto-1M dataset. These 50K samples include 25K reference-image/original-video pairs and \textbf{25K} reference-image/edited-video pairs, so edited videos only account for \textbf{3.3\%} of the total data. This 3.3\% video editing data is not used for main supervision, but only as data augmentation for cross-domain scenarios. Original-video/edited-video pairs are not used for training. We conduct an ablation study on whether to use Ditto-1M, as shown in Tab. \ref{video_editing} of the supplementary material B.

\section{Experiments}

\subsection{Experimental Setups}
\paragraph{\textbf{Implementation.}}  For a comprehensive verification, we train our model on both open-source text-to-video models Wan2.1-14B-T2V \cite{wan2025wan} and Wan2.2-14B-T2V. We adopt a two-stage training process: in the first stage, we fine-tune for 2,000 steps on a 200K image-personalization dataset with a batch size of 96, updating only the patch embedding and self-attention modules to acquire basic personalization while preserving the base model’s capabilities. In the second stage, we finetune for 12,000 steps on a 750K video-personalization dataset with a batch size of 64, freezing the cross-attention modules to maintain text-following ability.  We use the Adam optimizer with a learning rate of $1e-5$ in the training process. The total training cost is approximately 30,000 A100 GPU-hours. During training, the parameters of the reference branch in the Domain-MoT are initialized by copying those from the video branch, and the weight coefficient of the CCL Loss $\mathcal{L}_{\mathrm{C}}$ is set to $0.1$. During training, $K=4$, representing four domain attributes: real-world human, real-world object, background, and fantasy subject. Notably, domain attributes refer to the subject attributes in the generated video, rather than those in the reference images. Both the training and test sets use MLLM to annotate domain attributes. In addition, users can freely provide the corresponding domain attributes during their own inference.


\begin{table*}[t]
\centering
\caption{\textbf{Quantitative results}. The best scores are shown in \textbf{bold}, and the second-best are \underline{underlined}. \textit{DomainShuttle} significantly outperforms the baselines in text controllability and most subject consistency metrics.}
\setlength{\tabcolsep}{6pt}
\resizebox{1.0\textwidth}{!}{
\begin{tabular}{lccccccccc}
\toprule
\multirow{2}{*}{Method} 
& \multicolumn{2}{c}{Video Quality} 
& \multicolumn{1}{c}{Text Controllability} 
& \multicolumn{4}{c}{Cross-Domain Subject Consistency} 
& \multicolumn{2}{c}{In-Domain Subject Consistency} \\
\cmidrule(lr){2-3} 
\cmidrule(lr){4-4} 
\cmidrule(lr){5-8} 
\cmidrule(lr){9-10}
& AES$\uparrow$ 
& MS$\uparrow$ 
& GMEScore$\uparrow$ 
& NANO-CLIP$\uparrow$ 
& Qwen-CLIP$\uparrow$ 
& CD-Score$\uparrow$ 
& Qwen-Score$\uparrow$ 
& DINO-I$\uparrow$ 
& CLIP-I$\uparrow$ \\ 
\midrule
Kling 1.6 \cite{kling}              
& 0.515 & 0.965 & 0.596 & 0.621 & 0.640 & 0.725 & 0.771 & 0.401 & 0.672 \\ 
\midrule
VACE-Wan2.1-14B \cite{vace}         
& \textbf{0.517} & \underline{0.985} & 0.671 & 0.622 & 0.644 & 0.538 & 0.769 & 0.326 & 0.695 \\
MAGREF \cite{deng2026magref}        
& 0.491 & 0.964 & 0.678 & 0.618 & 0.638 & 0.499 & 0.705 & 0.312 & 0.685 \\
SkyReels-V3 \cite{skyreelsv3}       
& 0.481 & 0.920 & 0.656 & 0.593 & 0.616 & 0.493 & 0.681 & \textbf{0.407} & 0.673 \\
Phantom \cite{phantom}              
& 0.515 & 0.972 & 0.660 & 0.602 & 0.645 & 0.506 & 0.703 & 0.322 & \underline{0.701} \\
HuMo \cite{chen2025humo}            
& 0.479 & 0.981 & 0.663 & 0.609 & 0.636 & 0.495 & 0.681 & 0.317 & 0.682 \\
BindWeave \cite{li2026bindweave}    
& 0.450 & 0.963 & 0.617 & 0.598 & 0.612 & 0.510 & 0.629 & 0.317 & 0.681 \\ 
FFGO-Wan2.2-14B \cite{chen2025first} 
& 0.410 & 0.945 & 0.653 & 0.589 & 0.611 & 0.558 & 0.667 & 0.274 & 0.662 \\
VACE-Wan2.2-14B \cite{vace}         
& 0.480 & 0.974 & 0.685 & 0.606 & 0.622 & 0.546 & 0.679 & 0.303 & 0.679 \\
\midrule
Ours (Wan2.1-14B)                   
& 0.510 & 0.977 & \underline{0.689} & \underline{0.627} &  \underline{0.647}  & \underline{0.787}  & \underline{0.781} & \underline{0.405} & \textbf{0.703} \\ 
Ours (Wan2.2-14B)                   
& \underline{0.516} & \textbf{0.987} & \textbf{0.705} & \textbf{0.636} & \textbf{0.658} & \textbf{0.861}  & \textbf{0.829}  & 0.400 & 0.690  \\ 
\bottomrule
\end{tabular}}
\vspace{-10pt}
\label{tab:s2v_result}
\end{table*}

\paragraph{\textbf{Test Dataset.}}  We construct a test set consisting of $110$ in-domain samples and $110$ cross-domain samples. Among the 110 in-domain samples in the test set, 90 are from the open source OpenS2V-Eval\cite{yuan2025opens2v} data, while the remaining samples are self-constructed. The regular in-domain samples cover typical scenarios including multi-person, multi-object, human–object interactions, and background preservation. The cross-domain samples include real-to-fantasy transformations, fantasy-to-real transformations, and complex interaction cases between real subjects and fantasy characters within either real-world or fantasy domains.

\begin{figure*}[t]
    \includegraphics[width=0.98\linewidth]{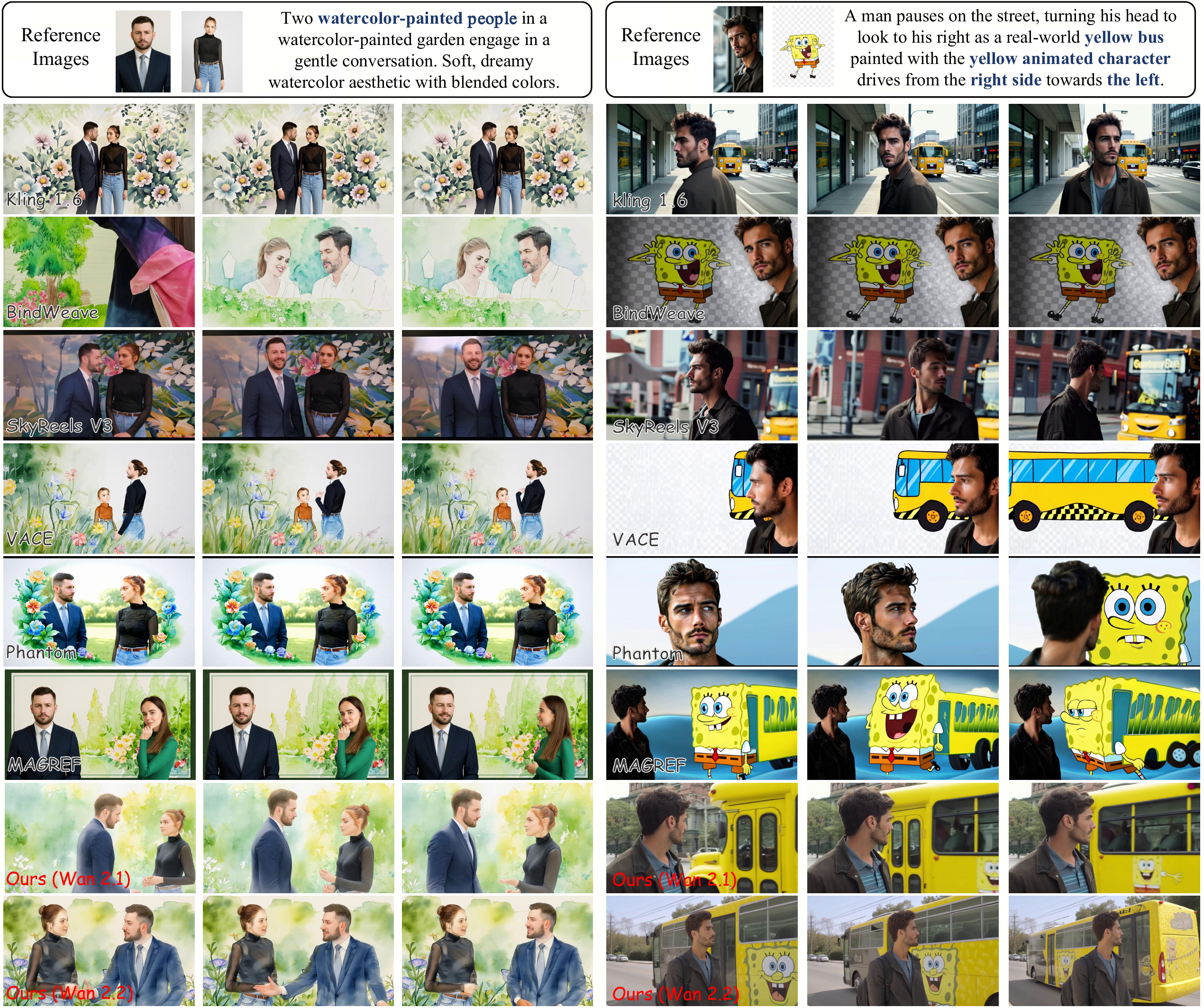}
    \caption{\textbf{Qualitative comparison} with existing methods. \textit{DomainShuttle} outperforms existing methods in cross-domain scenarios, which achieves flexible text controllability (\eg, the yellow bus printed with the character) and precisely preserves the features of the reference subject. }
    \label{fig:visual_case}
\end{figure*}

\begin{figure*}[t]
    \includegraphics[width=0.98\linewidth]{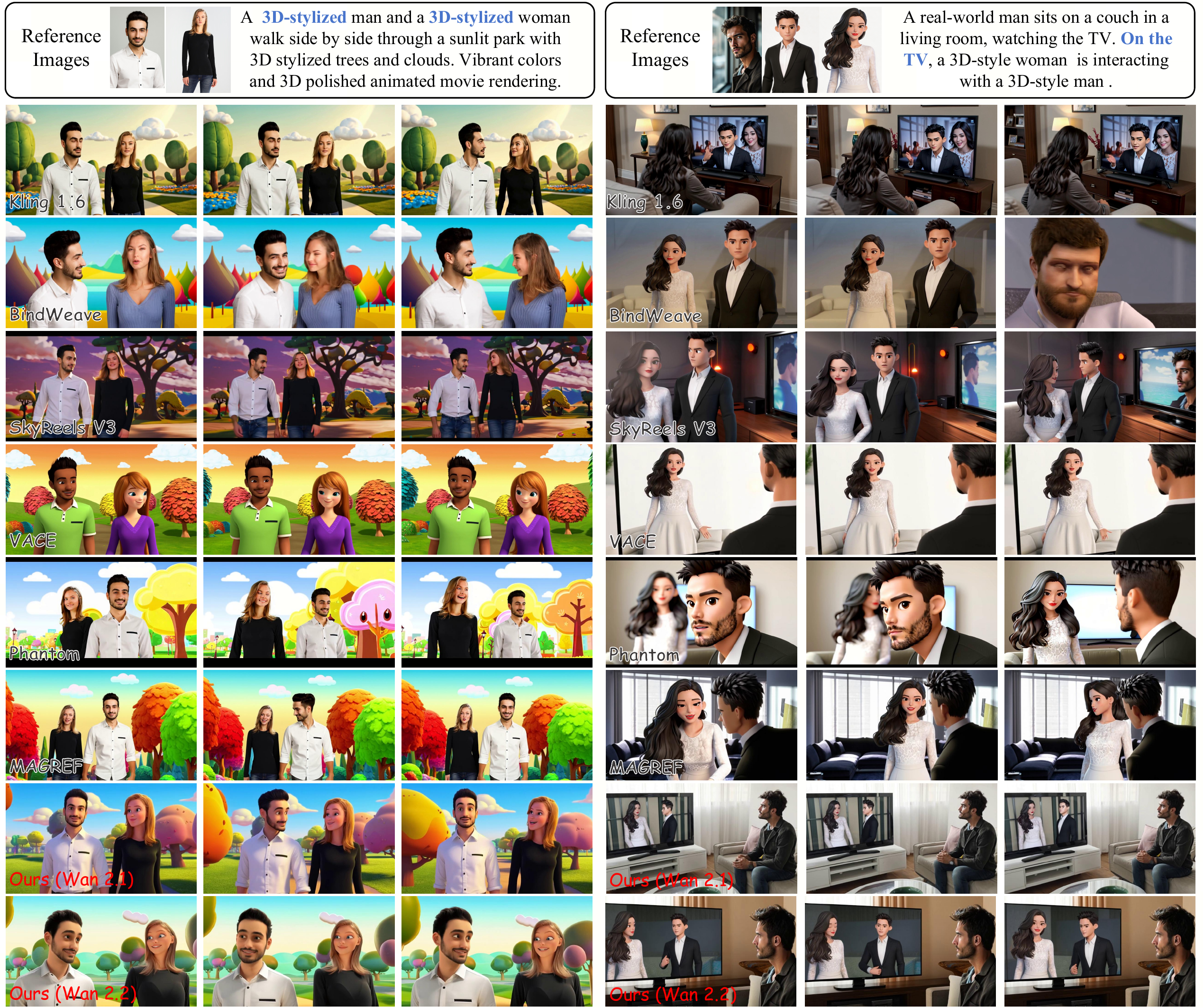}
    \caption{\textbf{More Qualitative comparison} with existing methods. \textit{DomainShuttle} further demonstrates strong performance in more challenging cross-domain scenarios. }
    \label{fig:visual_case_2}
\end{figure*}

\paragraph{\textbf{Evaluation Metrics.}} We evaluate the effectiveness of \textit{DomainShuttle} from three aspects. (1) Video Quality: The normalized Aesthetic Score (AES) and Motion Smoothness (MS) \cite{yuan2025opens2v} are used to evaluate the quality of videos generated by different methods. (2) Text Controllability: GMEScore\cite{gme} is used to evaluate text controllability. (3) Subject Consistency: We conduct evaluations under both In-Domain and Cross-Domain settings. For the In-Domain scenario, we adopt the standard DINO-I \cite{dinov2} and CLIP-I \cite{CLIP} metrics. Specifically, we first segment the subjects in videos and then compute subject-level similarity.

For the \textbf{cross-domain} scenario, we design four metrics to evaluate different methods: NANO-CLIP, Qwen-CLIP, Cross-Domain (CD) Score, and Qwen-Score. NANO-CLIP and Qwen-CLIP use Nano-Banana Pro\cite{nano_banana} and Qwen-Image-Edit\cite{wu2025qwenimage}, respectively, to generate cross-domain reference images, and then compute the CLIP similarity between these reference images and the generated videos. CD-Score and Qwen-Score evaluate each method using GPT-5.2\cite{gpt52} and the open-source Qwen3-VL-8B-Instruct\cite{bai2025qwen3}, respectively. NANO-CLIP and Qwen-CLIP follow the same evaluation pipeline, differing only in the image editing model. Similarly, CD-Score and Qwen-Score differ only in the MLLM used. The use of an open-source MLLM improves the reproducibility of the metrics, while cross-model evaluation further enhances the reliability of the assessment. More evaluation details are shown in Sec. \ref{Experiments_Results} of the supplementary materials.

\paragraph{\textbf{Baselines.}} We conduct quantitative analysis with SOTA methods to evaluate the effectiveness of our model. These methods are divided into three categories: (1) closed-source model Kling1.6 \cite{kling}; (2) Wan2.1-14B-based models: VACE \cite{vace}, MAGREF \cite{deng2026magref}, SkyReels-V3 \cite{skyreelsv3}, Phantom \cite{phantom}, HuMo \cite{chen2025humo}, and BindWeave \cite{li2026bindweave}; (3) Wan2.2-14B-based models: FFGO \cite{chen2025first} and VACE-Wan2.2 \cite{vace}.

\subsection{Main results}
\paragraph{\textbf{Quantitative Results.}} The quantitative results are shown in Tab. \ref{tab:s2v_result}. Compared with baselines, \textit{DomainShuttle} achieves the best performance in motion smoothness and text controllability. Our method significantly outperforms baselines in most subject consistency metrics, particularly with a significant $18.7\%$  improvement in cross-domain (CD) score. These quantitative experiments demonstrate that \textit{DomainShuttle} can generate high-quality videos with competitive text controllability and subject consistency in open-domain scenarios.

\paragraph{\textbf{Qualitative Results.}} Fig. \ref{fig:visual_case} and Fig. \ref{fig:visual_case_2} illustrate the qualitative results compared to existing methods. Our method demonstrates high subject consistency and text controllability. For real-world subjects in fantasy domains, such as watercolor and 3D animation domains (Fig. \ref{fig:visual_case} and Fig. \ref{fig:visual_case_2}, left side), our method preserves the inherent subject features while successfully following the style instructions, whereas existing methods either fail to follow the style guidance or lose subject consistency while following the style guidance. Our method also successfully achieves mapping fantasy subjects to real-world objects (Fig. \ref{fig:visual_case}, right side), while the baseline either fails to attach the fantastic subject to the bus or only generates the fantastic subject. For interactions between real-world and fantastic subjects (Fig. \ref{fig:visual_case_2}, right side), \textit{DomainShuttle} also achieves the best performance, while other methods fail to generate correct subject interactions. 

\subsection{Ablation Study}
To demonstrate the effectiveness of each essential module of \textit{DomainShuttle}, we conduct extensive ablation experiments. All ablation experiments are trained on the Wan2.2-14B-T2V with the same training steps.

\begin{table*}[t]
\centering
\caption{\textbf{Ablation Studies} of each essential module. }
\setlength\tabcolsep{10pt}
\resizebox{1.0\textwidth}{!}{%
\begin{tabular}{lcccccc}
\toprule
\multirow{2}{*}{ID} & \multirow{2}{*}{Setting} & Text Controllability  & \multicolumn{2}{c}{Cross-Domain Subject Consistency} & \multicolumn{2}{c}{In-Domain Subject Consistency} \\
 \cmidrule(lr){3-3} \cmidrule(lr){4-7} & &  GMEScore$\uparrow$ & NANO-CLIP$\uparrow$ & CD-Score$\uparrow$ & DINO-I $\uparrow$ & CLIP-I    $\uparrow$ \\

\midrule
0 & Naive Method       & 0.664 & 0.601 & 0.697 & 0.356 & 0.675  \\
1 & 0 + Dual Self-Attn   & 0.671 & 0.609 & 0.715 & 0.367 & 0.683  \\
2 & 0 + Domain-MoT       & 0.687 & 0.627 & 0.783 & \underline{0.396} & \textbf{0.697}  \\
\midrule
3 & 2 + VR-DualRoPE& \underline{0.691} & \underline{0.629} & \underline{0.813} & 0.394 & 0.688  \\
4 & 3 +  CCL            & \textbf{0.705} & \textbf{0.636} & \textbf{0.861} & \textbf{0.400} & \underline{0.690}  \\  
\bottomrule
\end{tabular}%
}
\vspace{-4pt}
\label{tab:ablation}
\end{table*}

\begin{figure*}[t]
    \includegraphics[width=0.98\linewidth]{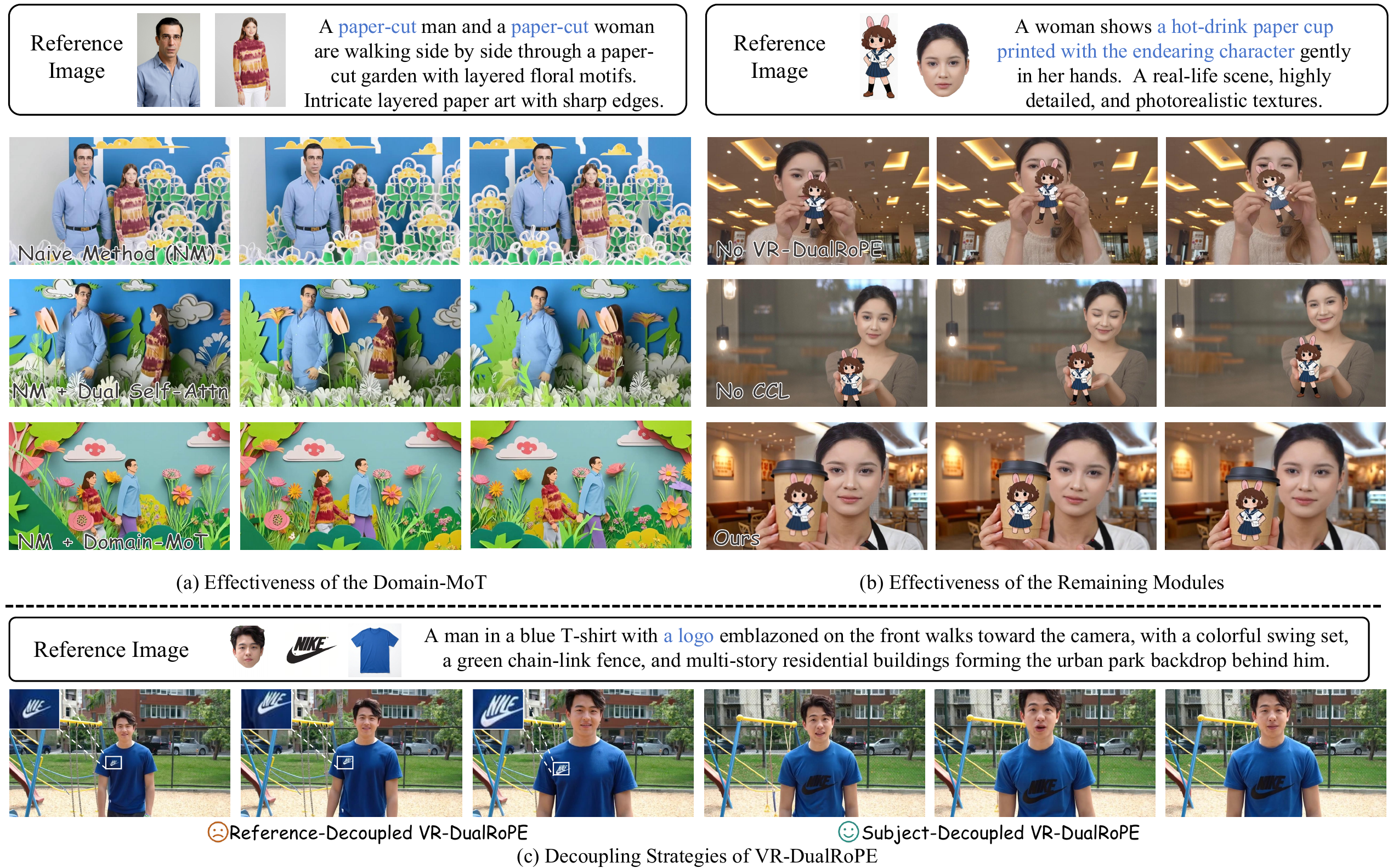}
    \caption{\textbf{Ablation Studies}. (a) Compared to Dual Self-Attn, Domain-MoT transfers all reference subjects into the fantasy domain. (b) Replacing VR-DualRoPE with naive RoPE causes incorrect subject interactions, while removing CCL induces direct copying from the reference. Using both yields the best results. (c) VR-DualRoPE uses a subject-decoupled offset to better bind these references than reference-decoupled offset.}
    \label{fig:ablation_fig}
\end{figure*}

\paragraph{\textbf{Effectiveness of the Domain-MoT.}}  Ablation results of Domain-MoT are shown in Tab. \ref{tab:ablation} and Fig. \ref{fig:ablation_fig} (a). The Naive Method refers to directly concatenating reference image tokens with video tokens without additional operations; Dual Self-Attn refers to introducing a dedicated attention mapping pathway for reference image tokens in self-attention. ID-0, ID-1, and ID-2 employ a naive RoPE scheme, in which the RoPE indices of the reference image tokens are concatenated to the video tokens along the temporal dimension. The Naive Method (ID-0) usually fails to transfer the reference subjects to the target fantasy domain. Dual Self-Attn (ID-1) partially improves domain transfer, but still cannot consistently transform all reference subjects. For instance, in  Fig. \ref{fig:ablation_fig} (a), only the woman is converted into the paper-cut style. In contrast, Domain-MoT (ID-2) successfully converts all humans into the target style. Quantitatively, Domain-MoT improves the CD-Score from 0.697 to 0.783 compared to the Naive Method.

\paragraph{\textbf{Effectiveness of the Remaining Modules.}} The ablation results of Video-Reference (VR) DualRoPE and CCL are shown in Tab. \ref{tab:ablation} and Fig. \ref{fig:ablation_fig} (b). Replacing VR-DualRoPE with the naive RoPE scheme causes the reference image to be treated as a video frame in the RoPE space, leading to incorrect subject interactions and weaker editability. Without VR-DualRoPE, the animated character appears at an incorrect spatial position and fails to attach to the paper cup held by the person, as shown in Fig. \ref{fig:ablation_fig} (b). Removing CCL causes the model to directly copy the object from the reference image, indicating that CCL facilitates learning more precise representations of the reference subject. CCL mainly improves \textbf{controllability} in cross-domain scenarios, rather than fidelity. As shown in Tab. \ref{tab:ablation} of the main submission, CCL improves fidelity by only 0.3\% (CLIP) and 1.5\% (DINO), but significantly improves CD-Score by \textbf{5.9\%}, supporting this point. Combining VR-DualRoPE and CCL, \textit{DomainShuttle} (ID-4) achieves the best performance.


\paragraph{\textbf{Decoupling Strategies of VR-DualRoPE.}} In some open-domain scenarios, multiple reference images may represent different attributes of the same subject. To address this, Video-Reference DualRoPE adopts a strategy in which multiple reference images of the same subject are offset only along the width dimension in RoPE, rather than applying offsets along both height and width for different reference images by default. As shown in Fig. \ref{fig:ablation_fig} (c), compared with the reference-decoupled strategy, the subject-decoupled strategy better binds multiple reference images of the same subject.

\subsection{Human Preference Evaluation}

\begin{figure*}[t]
    \includegraphics[width=0.98\linewidth]{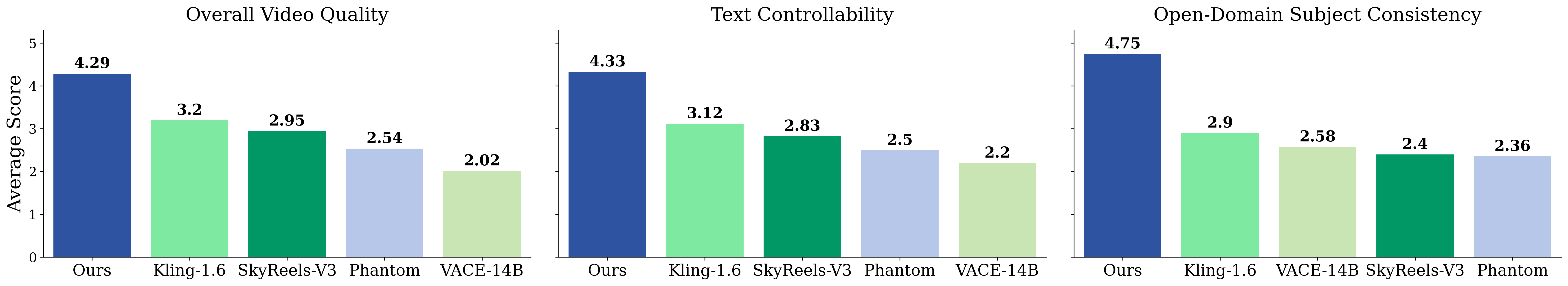}
    \caption{\textbf{Human preference evaluation}. \textit{DomainShuttle} significantly outperforms baselines in video quality, text controllability, and open-domain subject consistency.}
    \label{fig:user_study}
\end{figure*}

We invite $40$ volunteers to conduct a human preference evaluation comparing \textit{DomainShuttle} with well-performed methods. Each person ranks $20$ randomly selected videos in three aspects: video quality, text controllability, and open-domain subject consistency. Distinct scores from 5 (best) to 1 (worst) are assigned to different methods without ties. As shown in Fig. \ref{fig:user_study}, our method achieves superior performance in all metrics, showing significant advantages in open-domain subject consistency, validating the effectiveness of \textit{DomainShuttle}.

\section{Conclusion}

In this paper, we propose \textit{DomainShuttle}, a novel architecture designed to achieve high fidelity and generative flexibility for open-domain video personalization. \textit{DomainShuttle} introduces Domain-MoT to decouple video and reference features for domain-aware reference modeling, Video-Reference DualRoPE to separate the RoPE space of reference images for fine-grained subject-level spatial modeling, and Cross-Pair Consistent Loss to accurately extract intrinsic subject representations. Extensive experiments demonstrate that our model achieves competitive performance in both cross-domain and in-domain scenarios, effectively optimizing both subject consistency and generative flexibility while exhibiting strong generalization across various complex applications.

\bibliographystyle{unsrtnat}
\bibliography{reference}

\clearpage
\beginappendix

In the supplementary materials, we provide the construction of the training set in \cref{TrainingSet}, and present more experimental setup and results in \cref{Experiments_Results}. We strongly recommend viewing the static HTML files in the supplementary materials for a direct and clear demonstration of the unique capabilities of our model and significant improvements over previous methods.

\section{Construction of the Training Dataset}
\label{TrainingSet}

Our training datasets are all open-source datasets, primarily including open-source image personalization datasets and open-source video personalization datasets. Specifically, image personalization datasets consist of four parts: UNO \cite{UNO}, Echo-4o \cite{echo-4o}, MUSAR \cite{musar}, and Nano-Consistent-150K \cite{echo-4o}. UNO and Nano-Consistent-150K are single-subject datasets, and the remaining datasets are multi-subject datasets. We primarily filter the datasets based on aesthetic quality and personalization quality scores using an MLM-based evaluation system, ultimately obtaining 200K high-quality image datasets to enable text-to-video models with basic personalization ability. The video personalization datasets we use are also all open-source datasets; the number of samples after filtering for each dataset is shown in Tab. \ref{tab:datasets}.

\begin{table}[ht]
  \caption{Detailed description of the training dataset.}
  \label{tab:datasets}
  \centering
  \resizebox{0.85\textwidth}{!}{
  \begin{tabular}{@{}cccc@{}}
    \toprule
    Dataset Name & Filtered Dataset Size & Modality & Category  \\
    \midrule
    UNO \cite{UNO} & 50K & Image & Single Subject  \\
    Nano-Consistent-150K \cite{echo-4o} & 60K & Image & Single Subject \\
    Echo-4o\cite{echo-4o}  & 60K & Image & Multi Subject \\
    MUSAR\cite{musar} & 30K & Image & Multi Subject \\
    Total Dataset & 200K & Image & - \\
    \midrule
    Phantom-Data\cite{phantom-data} & 400K  & Video & Single \& Multi Subject \\
    Opens2v\cite{yuan2025opens2v} & 300K & Video & Multi Subject \\
    Ditto-1M\cite{ditto1m} & 50K & Video & Single \& Multi Subject \\
    Total Dataset & 750K & Video & - \\
  \bottomrule
  \end{tabular}}
\end{table}

\begin{figure*}[h]
    \includegraphics[width=0.98\linewidth]{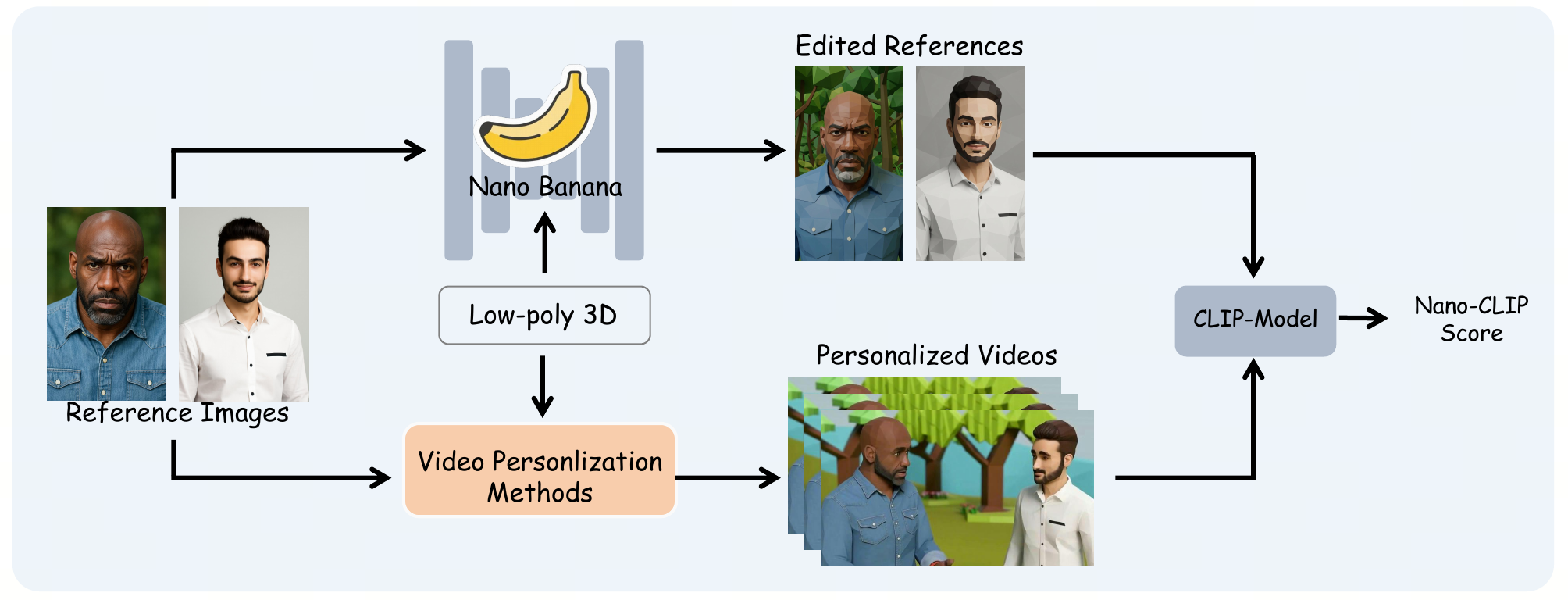}
    \caption{The calculation process of the Nano-CLIP metric. Given reference images and a domain transformation prompt, Nano Banana Pro first generates edited reference images. Each video personalization method then generates videos conditioned on the prompt. Finally, CLIP computes the cosine similarity between the generated frames and the edited references. The average similarity across frames is used as the Nano-CLIP score. Qwen-CLIP follows the same pipeline, except that the image editing model is replaced with Qwen-Image-Edit.
}
    \label{fig:nano}
\end{figure*}

\begin{table*}[!t]
\centering
\caption{MLLM prompt for Cross-Domain (CD) Score and Qwen-Score.}
\label{tab:cd score}
\resizebox{1.0\textwidth}{!}{%
\begin{tabular}{p{0.95\textwidth}}
\toprule
Guidelines: You are an experienced AI evaluator for cross-domain subject-driven text-to-video generation. You will receive two images and a reference caption.\\
\midrule
Reference Image 1: a reference subject. \\
Image 2 from Generated Video: the subject in a cross-domain scenario. For example, a person from the real world domain is transformed into a fantasy character, or an anime character is transformed into a real-world doll. \\
Reference Caption:  describes the intended subject and scene.
\\ \\
Task: \\
Your task is to measure the performance of transforming reference image 1 to image 2 based on the reference caption and the intrinsic features of reference subjects (\eg, shape, color, texture, and appearance). \\

\midrule

Scoring Rules: \\
1. Assign a score between 1 and 5. \\
2. Metric: \\
         \hspace{1em}  \textbf{5}: Achieving good cross-domain transformation while preserving the most features in reference image 1; \\
         \hspace{1em} \textbf{4}: Achieving cross-domain transformation of key features in reference image 1, but some negative and non-critical feature transformations have flaws. \\
        \hspace{1em} \textbf{3}: Achieving cross-domain transformation for part key features (such as human body), but domain transformation is not achieved for the remaining key and minor features. \\
         \hspace{1em} \textbf{2}: Only achieving the transformation of some negative and non-critical features in reference image 1, such as background and minor decorative elements, while the transformation of key features (e.g., human face) fails. \\
         \hspace{1em} \textbf{1}: Almost completely copies and pastes the features of reference image 1 without any cross-domain transformation. Alternatively, the cross-domain transformation result completely discards the features of reference image 1. \\ \\
3. Higher scores correspond to higher cross-domain consistency. \\
Strict Output Requirement: Return only the numerical score (no explanations, text descriptions, or additional comments).  \\
 \bottomrule
\end{tabular}%
}
\end{table*}

\begin{table*}[!t]
\centering
\caption{Guidelines of human preference evaluation.}
\label{tab:user_study}
\resizebox{1.0\textwidth}{!}{%
\begin{tabular}{p{0.95\textwidth}}
\toprule
Guidelines: Open-Domain subject-driven video (S2V)  personalization is an important downstream task of text-based video generation, aiming to generate videos based on user-provided images and prompts.  Open domain S2V mainly involves two scenarios: in-domain (\eg, multi-human, multi-object, and human-object interaction) scenarios and cross-domain (\eg, real-world subject to fantastic domain, fantastic subject to real-world domain, and the interaction of the real-world and fantastic subjects) scenarios. Please watch the following randomly selected videos, reference images, and prompts. Compare their effects and evaluate the generated video based on three metrics: \\

\midrule
1. Overall Video Quality: Comprehensively evaluate the overall quality of the generated videos from three aspects: aesthetic quality, the smoothness of subject motions (avoiding static or frozen subjects and frame discontinuities), and the naturalness of color, texture, and saturation. \\

 \\
 
2. Evaluate text controllability based on the consistency between the generated video and the input text description (\eg, corresponding real-world or fantastic domain descriptions, stylistic attributes, and subject interaction alignment). \\

\\

3. Open-Domain Subject Consistency: Evaluate subject consistency based on the similarity between the generated subject and the subject of the reference images. In in-domain scenarios, the best methods require retaining the reference subject features as much as possible. In cross-domain scenarios, the best methods should preserve the intrinsic features of the subject (\eg, hairstyle, skin color, and clothing) while allowing subject-irrelevant properties (\eg, lighting, style, and domain attributes) to vary flexibly according to the text prompt.  \\
\midrule

Please rank these methods across these three metrics. \\
 \bottomrule
\end{tabular}%
}
\end{table*}

\section{More Experiments Results}

\label{Experiments_Results}

\subsection{Implementation Details}

\textit{DomainShuttle} utilizes the default settings for inference for both Wan2.1 and Wan2.2, using 50 sampling steps on Wan2.1 and 40 steps on Wan2.2. The classifier-free guidance scale is set to 3 in Wan2.1. In Wan2.2, the high-noise classifier-free guidance scale is set to 4, while the low-noise guidance scale is set to 3. All flow shift parameters are set to 5.

\subsection{Evaluation Details}

\paragraph{\textbf{Evaluation Dataset.}} The 110 in-domain testset includes 90 \textbf{open-source} cases from OpenS2V-Eval \cite{yuan2025opens2v}, with 30 each for multi-human, multi-object, and Hard\_dev cases. Because some methods support at most four subjects, we use the first four subjects in each Hard\_dev case as references. Additionally, the in-domain test set includes 20 human–object interaction cases. For the cross-domain evaluation dataset, we construct 40 multi-subject samples for real-world subjects in fantasy domains, 40 multi-subject samples to map the fantasy subjects to the real-world objects, and 30 multi-subject samples for interactions between real-world and fantastic subjects.

\paragraph{\textbf{Evaluation Metrics.}} For all subject similarity metrics, we uniformly sample 16 frames from the videos generated by each method, compute the score for each frame, and take the average score as the final result.

The evaluation process of Nano-CLIP is shown in Fig. \ref{fig:nano}. For the Nano-CLIP metric, we first use Nano Banana Pro \cite{nano_banana} to generate the edited reference images based on the reference images and the domain transformation prompt. Next, guided by the prompt, each video personalization method generates videos of the reference images. Finally, CLIP is used to calculate the cosine similarity between each frame of the generated videos and the reference images edited by Nano Banana. The average similarity across all frames is used as the Nano-CLIP score for these methods.

The Cross-Domain Score leverages the comprehensive understanding capability of the multimodal large language model (\ie, GPT-5.2) to thoroughly evaluate the intrinsic subject consistency of different methods in cross-domain scenarios. Detailed evaluation instructions are provided in Tab. \ref{tab:cd score}. We average all scores and apply normalization to obtain the final results.

\subsection{Evaluation Criteria for Human Preference Evaluation}

Each volunteer is presented with 20 randomly selected open-domain videos for our method and four strong baselines, resulting in 100 videos in total. For each group, participants are required to rank the videos under every metric, assigning scores from 5 (best) to 1 (worst). Ties are not allowed in any ranking for any metric. The instructions provided to the participants are shown in Tab. \ref{tab:user_study}.

\subsection{More Qualitative Comparisons.}

\begin{figure*}[!t]
    \includegraphics[width=0.98\linewidth]{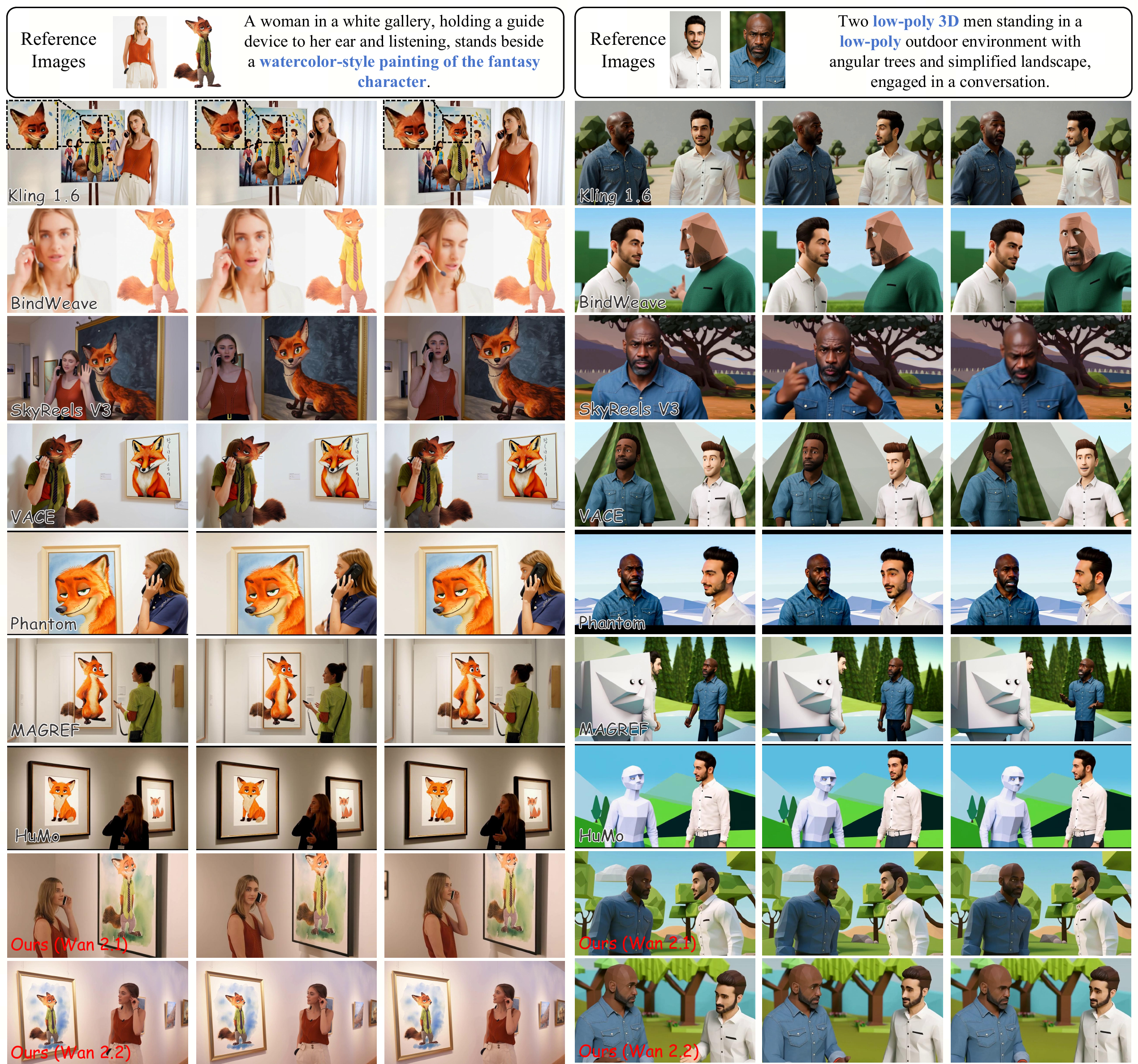}
    \caption{\textbf{More qualitative comparison} with existing methods. For real–fantasy subject interactions, \textit{DomainShuttle} successfully generates interactions between the woman and the painting of the fantasy character. Among existing methods, the commercial model Kling-1.6 performs comparatively well, but the painted character blinks, which contradicts the static nature of the painting. For real-to-fantasy scenario, our method accurately maps real-world subjects to the low-poly 3D domain. In contrast, existing methods either fail to transfer subjects or lose key subject features after transfer.}
    \label{fig:visual_case_6}
\end{figure*}

We present more qualitative comparisons between \textit{DomainShuttle} and the baselines in the supplementary materials, as shown in Fig. \ref{fig:visual_case_5} and  Fig. \ref{fig:visual_case_6}. Our method outperforms existing methods across all three scenarios: mapping real-world subjects to fantasy domain, mapping fantasy subjects to the real world, and interactions between fantasy and real-world subjects. These qualitative comparisons demonstrate that \textit{DomainShuttle}  can achieve flexible text controllability and precisely preserve the intrinsic features of the reference subject.

\subsection{More Ablation Study}

In addition, we conducted an ablation study on whether to use Ditto-1M, as shown in the Tab. \ref{video_editing}. Without Ditto-1M,  our method remains effective and still achieves cross-domain SOTA performance, improving the CD-Score by 13.5\% over baselines (0.725 of Kling 1.6).

\begin{table*}[t]
\centering
\caption{\textbf{Ablation Studies} of Ditto 1M. }
\renewcommand{\arraystretch}{1.15}
\small
\begin{tabularx}{\textwidth}{l *{4}{>{\centering\arraybackslash}X}}
\toprule
Method (Wan 2.2) & NANO-CLIP $\uparrow$ & CD-Score $\uparrow$ & DINO-I $\uparrow$ & CLIP-I $\uparrow$ \\
\midrule
w/o Ditto 1M & 0.631 & 0.823 & \textbf{0.432} & \textbf{0.701} \\
w Ditto 1M   & \textbf{0.636} & \textbf{0.861} & 0.400 & 0.690 \\
\bottomrule
\end{tabularx}
\vspace{-6pt}
\label{video_editing}
\end{table*}

\begin{figure*}[!t]
    \includegraphics[width=0.98\linewidth]{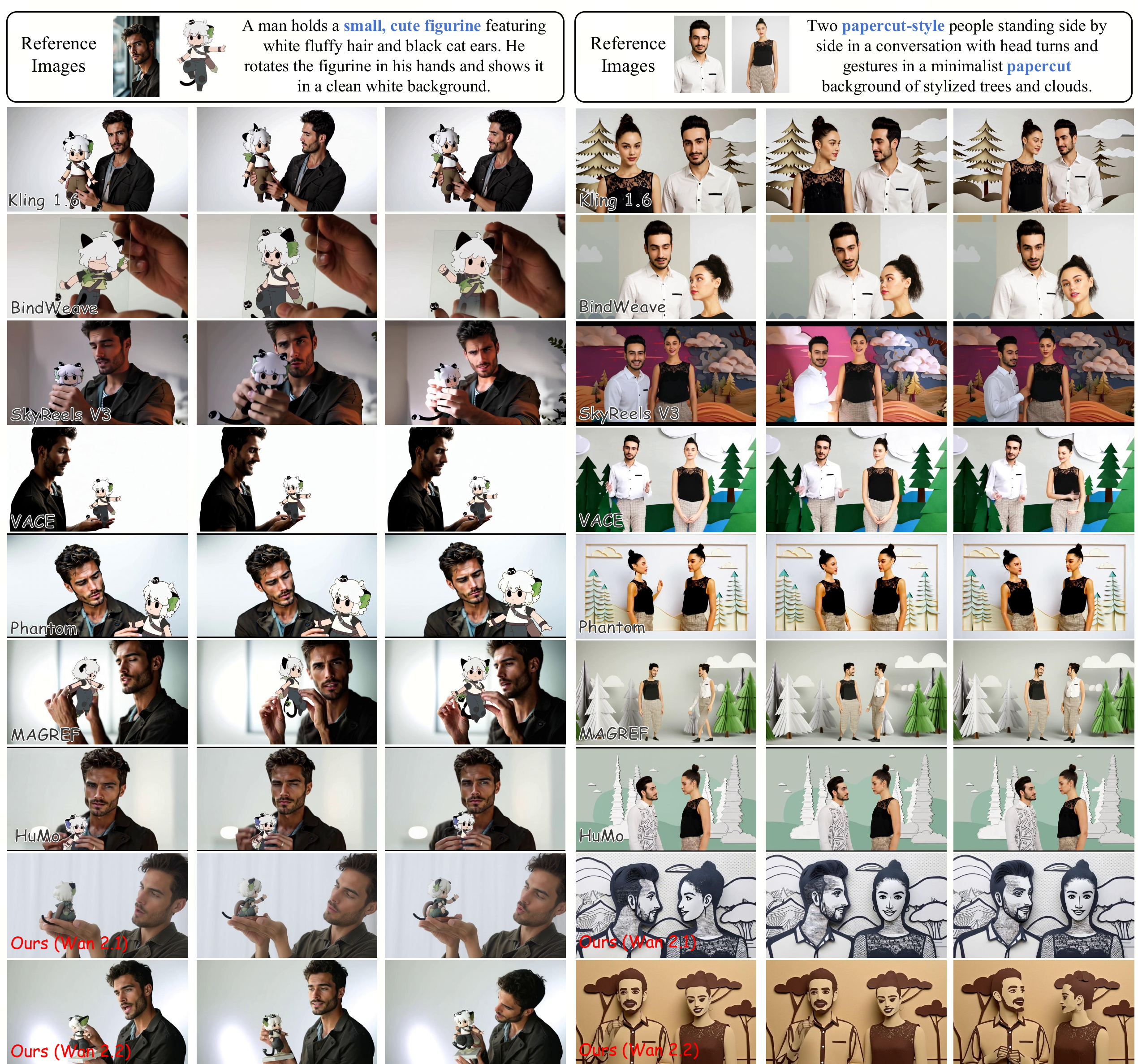}
    \caption{\textbf{More qualitative comparison} with existing methods. For mapping fantasy-domain subjects to real-world subjects, \textit{DomainShuttle} successfully converts the fantasy character into a real-world small figurine, outperforming existing methods. For real-to-fantasy, \textit{DomainShuttle} transforms real-world subjects into the corresponding paper-cut fantasy domain, whereas existing methods fail to achieve this conversion.
}
    \label{fig:visual_case_5}
\end{figure*}

\end{document}